\documentclass[letterpaper, 10pt, conference]{ieeeconf} 
\IEEEoverridecommandlockouts
\usepackage{moreverb,url}

\usepackage{epsfig}
\usepackage{graphicx}
\usepackage{amsmath}
\usepackage{amssymb}
\usepackage[pagebackref=true,breaklinks=true,letterpaper=true,colorlinks,bookmarks=false]{hyperref}
\usepackage{amsfonts}
\usepackage{algorithmic}
\usepackage{graphicx}
\usepackage{textcomp}
\usepackage{xcolor}
\usepackage{array}
\usepackage{mathtools}
\usepackage{tensor}
\usepackage{float} 
\usepackage{subfigure}
\usepackage[frozencache,cachedir=.]{minted}

\begin{document}

\title{Underwater inspection and intervention dataset}

\author{Tomasz {\L}uczy{\'n}ski\textsuperscript{a}, Jonatan Scharff Willners\textsuperscript{a}, Elizabeth Vargas\textsuperscript{a}, Joshua Roe\textsuperscript{a},\\ Shida Xu\textsuperscript{a}, Yu Cao\textsuperscript{b}, Yvan Petillot\textsuperscript{a} and Sen Wang\textsuperscript{a} \\ 
\textsuperscript{a }\textit {Institute of Sensors, Signals and Systems, Heriot-Watt University, UK, t.luczynski@hw.ac.uk; s.wang@hw.ac.uk}\\
\textsuperscript{b }\textit {The University of Edinburgh, UK}
}

\maketitle

\begin{abstract}
	This paper presents a novel dataset for the development of visual navigation and simultaneous localisation and mapping (SLAM) algorithms as well as for underwater intervention tasks. It differs from existing datasets as it contains ground truth for the vehicle's position captured by an underwater motion tracking system. The dataset contains distortion-free and rectified stereo images along with the calibration parameters of the stereo camera setup. Furthermore, the experiments were performed and recorded in a controlled environment, where current and waves could be generated allowing the dataset to cover a wide range of conditions - from calm water to waves and currents of significant strength. 
\end{abstract}


\thispagestyle{plain} 
\pagestyle{plain} 

\section{Introduction}
\label{sec:introduction}
In general, marine robotics deals with the same problems as other forms of field robotics: vehicle navigation, mapping, scene understanding and intervention in the state of the environment. However, the unique and hostile underwater conditions make this field of research unlike any other. 
One of the greatest obstacles on the way to the rapid development of systems and methods is the necessity for field experiments. These are not only time consuming and expensive but also there is usually little ground truth data to evaluate the progress. Furthermore, the conditions can vary significantly over time and the proposed approach often needs to be evaluated multiple times in various locations or at different time. 
Arguably, underwater vehicles often move in open spaces, where accurate localization is not essential. However, moving through these open spaces is usually on the way to a location that needs to be inspected or where some other task needs to be performed. 
Upon arrival to a structure or region of interest, accurate localization becomes essential. Apart from the safety of the vehicle during mission, if a robot cannot georeference an observed feature with a high enough reliability, the data may lose much of its value.  However, opposed to surface and aerial robots, sub-sea is a GPS denied environment and the robot, therefore, has to rely on different means for localisation. Various SLAM algorithms are commonly used to address this problem. In some cases, dead reckoning can be used, however, it is based on the integration of sensor data containing noise and will have an unbounded drift. Visual SLAM is able to use features in the environment to bound this error.
Optical cameras, with their low cost, high update rate and a data format that can be easily interpreted by humans make them the most common sensor, both for operators and SLAM. 
For that reason, this paper along with the provided dataset focuses on the visual data in the form of two video streams from a calibrated stereo setup.

There are datasets available online, e.g. \cite{girona} that provide sensor data from field experiments that can be used for testing SLAM algorithms, but there is no access to ground truth for validation. Therefore, the results can only be assessed qualitatively or by comparison to other approaches, but quantitative evaluation is limited to distinct features at re-visited places, distances between markers (if present) etc. Furthermore, this dataset provides a long sequence of data, but there is relatively little variation in environmental conditions. There is therefore a risk of building an approach fitted to these data but not generalising well. 
Finally, there is no data available that could be used to support the development of manipulation algorithms or any other intervention missions. This can seem obvious as the task inherently requires interaction and feedback from the environment, that cannot be captured in a recorded dataset.
However, with the dataset presented in this paper, we aim to rectify the issues typical for the visual navigation/SLAM datasets, but also propose data samples that can be used to support the development of manipulation tasks. It comprises rectified stereo images supported by the ground truth motion data collected by an underwater motion capture system. Multiple recordings in changing conditions are included, to provide data samples for smooth motion as well as the vehicle moving in increasingly harsh conditions, against current and waves.

\section{Sensors and facilities}
In this section, we present the remotely operated vehicle (ROV) used and the sensors it was equipped with, along with the environment the data was collected in.

\subsection{ROV}
\begin{figure}[h]
	\centering
	\includegraphics[width=.8\linewidth]{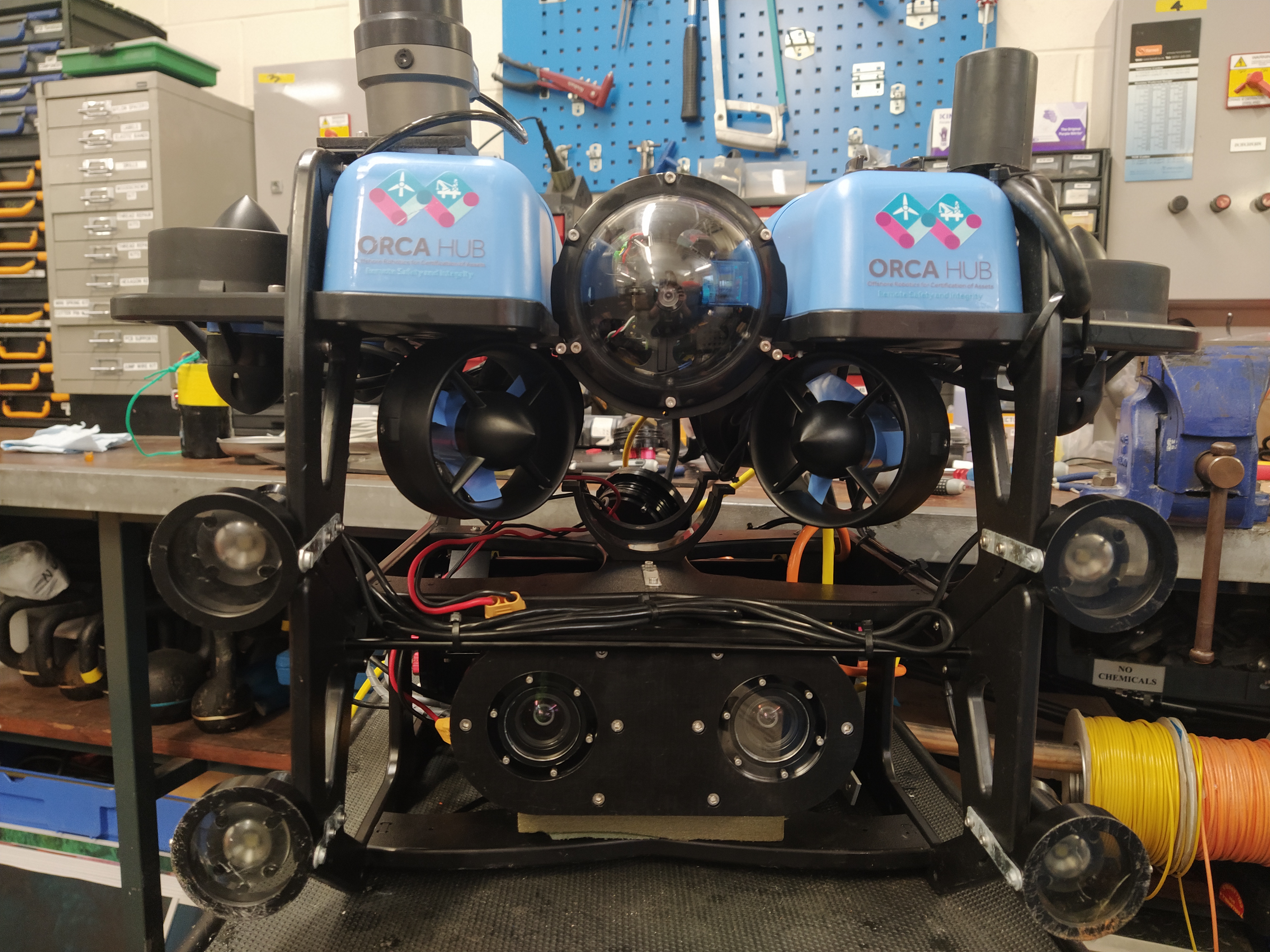}
	\caption{BloueROV2 heavy - a vehicle used in the experiments. The skid attached to the vehicle is equipped with the underwater stereo camera.}
	\label{fig:bluerov}
\end{figure}

During the experiments, a small BlueROV2 (Fig. \ref{fig:bluerov}, \cite{bluerov}) vehicle was used. It was assembled in the "heavy" configuration with 8 thrusters and a payload skid for sensors. Despite its small size (45 cm wide, 57 cm long and 40 cm tall) it is capable of operating in the presence of current and waves. Furthermore, it allowed for a significant range of motion in the tank, where the experiments were recorded. A bigger vehicle, even though more similar to those used offshore, would not be able to cover the same range of motion and would not fit the scale of this experiment.

\subsection{Stereo images}
\begin{figure*}[h]
	\centering
	\includegraphics[width=.8\linewidth]{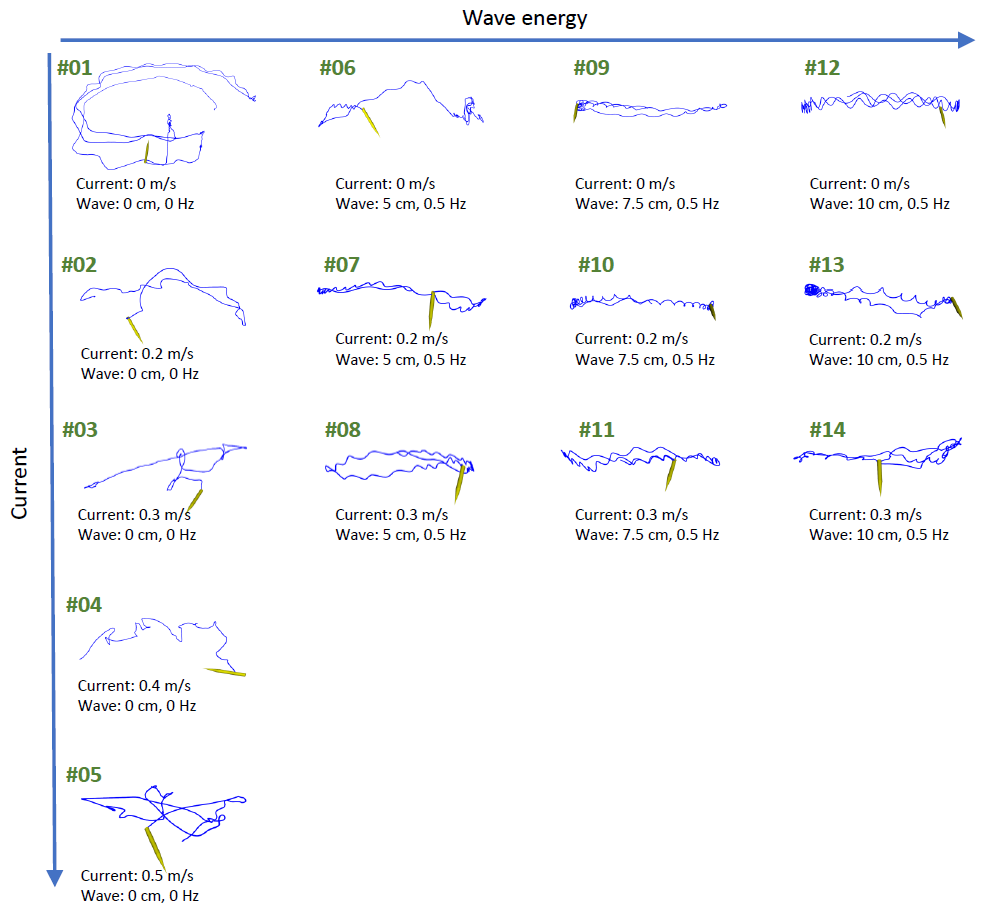}
	\caption{Vehicle's paths registered in the dataset. Yellow arrow shows the pose of the vehicle at the end of the sequence, pointing towards the structure. Sequence numbers, e.g. $\#01$, correspond to the names of directories in the dataset.}
	\label{fig:paths}
\end{figure*}
The main sensor used in the data collection was an underwater RGB stereo camera. The cameras were set in a master-slave configuration: whenever the master camera starts the exposure, it automatically triggers the slave camera. This allowed for registering images simultaneously. A USB 3.1 interface was used to minimize any delays caused by the communication layer.
When recording images underwater, the camera calibration also needs to be considered carefully. The water-glass-air interface introduces refraction-based distortions that, in the general case, are not trivial and invalidate the single viewpoint assumption, thus the pinhole camera model shall not be used \cite{TreibnitzPAMIcalib}. Our system was designed and built to comply with the Pinax camera model and calibration \cite{pinax}. There are multiple advantages to this approach, with the most important for this application being that the pinhole camera model can be successfully used.
Therefore any computer vision algorithms requiring the single viewpoint can be applied to these images, e.g. standard block matching algorithm for the stereo 3D reconstruction can be applied without any intermediate processing. The images in the dataset are already distortion-free and rectified.
Another point of concern, when working with underwater images, is the image degradation caused by light attenuation and scattering. Since the experiments were performed in a tank with very clear water, this is of lesser concern. However, an underwater white balance algorithm \cite{BiancoNeumannUW_WB} was applied, to ensure that the appearance of the registered scene is consistent, regardless of the distance and viewing angle.

The camera parameters characterising the system are presented here:

\begin{minted}[
	gobble=1,
	]{yaml}
	frame_id: left_camera
	framerate: 30
	width: 612
	height: 512
	camera_matrix: [655, 0, 306;
	                  0, 655, 256;
	                  0, 0, 1]
	dist_coeff: [0,0,0,0,0]
	projection_matrix: [655, 0, 306, 0;
	                      0, 655, 256, 0;
	                      0, 0, 1, 0]
	
	----------------------
	
	frame_id: left_camera
	framerate: 30
	width: 612
	height: 512
	camera_matrix: [655, 0, 306;
	                  0, 655, 256;
	                  0, 0, 1]
	dist_coeff: [0,0,0,0,0]
	projection_matrix: [655, 0, 306, -78.892;
	                      0, 655, 256, 0;
	                      0, 0, 1, 0]
\end{minted}

\subsection{Motion tracking system}
As discussed earlier, there are some underwater vision datasets available, e.g. \cite{girona,caddy_data}. However, from the perspective of the development of an autonomous underwater system, the main drawback of these datasets, or indeed almost any other that can be collected offshore, is the lack of the ground truth for the vehicle movement. For the dataset presented here, the vehicle was equipped with markers that were tracked by the Qualisys motion tracking system. It provides external position reference for the underwater vehicle. 
The system used, consisted of six underwater motion tracking cameras, a central processing unit and a high-speed Ethernet connection to the vehicle.
Several reflective balls were attached to each side of the underwater vehicle to allow at least three of them to be visible to the tracking system at any time.  
The central processing unit runs the propriety motion tracking software provided by the vendor.
Before the experiment, a T-shaped referenced object with four reflective balls was used the calibrate the system. Since the precise geometry of this T-shape reference object is known, a highly accurate position measurement within $1$ mm uncertainty residual can be achieved. The tracking data as well as the sensor data were recorded using ROS (\cite{ROS}).

\begin{figure}
	\centering  
		\includegraphics[width=0.49\linewidth]{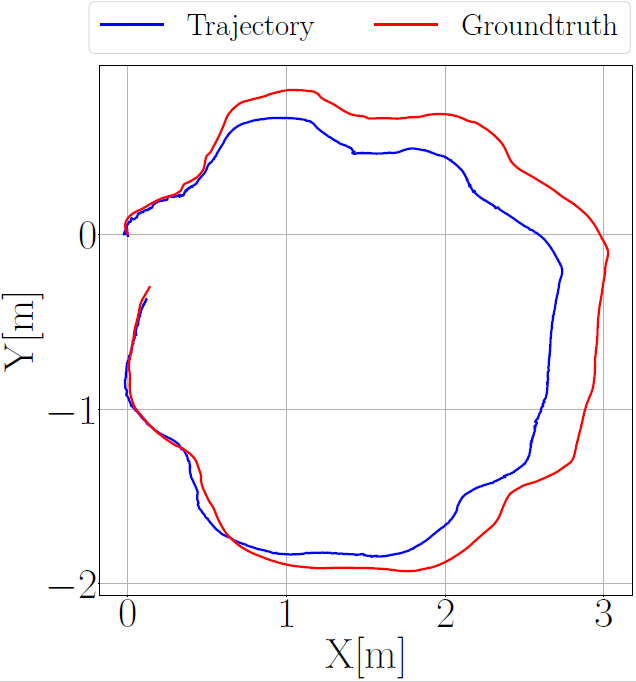}
		\includegraphics[width=0.49\linewidth]{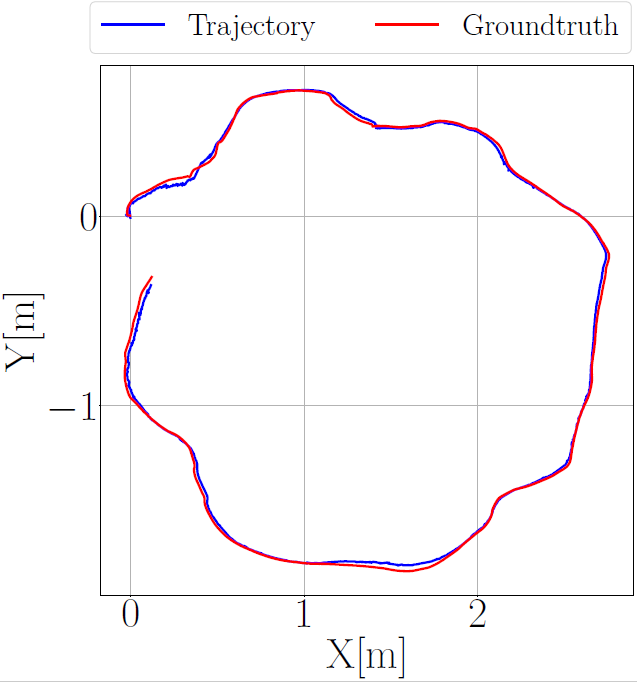}
	\caption{The trajectories before and after calibration.}
	\label{fig:qsys_cam_calib}
\end{figure}

The reference frame tracked by the Qualisys system and the camera frame are both fixed on the vehicle but differ from each other. To utilize the data recorded with the motion tracking system, the relative pose between these two must be found. To this end, a data sequence was recorded, where the vehicle was moving slowly and smoothly, to enable the best possible conditions for the visual SLAM algorithm. In this case, ORB-SLAM3 was used \cite{orbslam3}. This way, two odometry models are obtained: Qualisys, tracking the markers on the ROV, and visual SLAM, tracking the camera pose. Ideally, knowing the relative pose between these two frames, both trajectories should overlap.
For both the Qualisys and camera, the trajectories were recorded. $T_{c0}$ denotes the first camera frame and $T_{q0}$ the corresponding Qualisys frame. Both were saved and all later poses $T_{ci}$ and $T_{qi}$ in the recording were represented as the transformation from the initial to the current frame $\prescript{c0}{}T_{ci}={T_{c0}}^{-1}T_{ci}$ and $\prescript{q0}{}T_{qi}={T_{q0}}^{-1}T_{qi}$. The visualization of both paths is shown in Figure \ref{fig:qsys_cam_calib}.

The goal is to find a transformation from the Qualisys frame to the camera frame, denoted as $\prescript{q}{}T_{c}$. 
Knowing it, the pose registered $\prescript{q0}{}T_{qi}$ can be used to find the camera pose $\prescript{c0}{}T_{ci}^{quali}$:
\begin{equation}
	\prescript{c0}{}T_{ci}^{quali} = {\prescript{q}{}T_{c}}^{-1} \prescript{q0}{}T_{qi} \prescript{q}{}T_{c}
	\label{euq1}
\end{equation}
\\
If $\prescript{q}{}T_{c}$ and estimation of pose are ideal $\prescript{c0}{}T_{ci}=\prescript{c0}{}T_{ci}^{quali}$ should be true. This forms an optimization problem, that was solved to find the $\prescript{q}{}T_{c}$, with the cost function defined as \ref{cost}.
\begin{equation}
	\mathop{\arg\min}_{\prescript{q}{}T_{c}} \ \ \|  {\prescript{c0}{}T_{ci}}^{-1}p_{0}- {({\prescript{q}{}T_{c}}^{-1} \prescript{q0}{}T_{qi} \prescript{q}{}T_{c})}^{-1}p_{0}\|^2 \label{cost}
\end{equation}
This yielded the result:
\begin{minted}[
	gobble=1,
	]{yaml}
	translation: [-0.1326, 0.0059, 0.0018]
	rotation: [0.9979, -0.0143,-0.0632;
               0.0178, 0.9984, 0.0541;
               0.0624, -0.0551, 0.9965]
    
\end{minted}
Both paths aligned with the obtained transformation are presented in Figure \ref{fig:qsys_cam_calib}.

\subsection{Wave and current generation}
The data was recorded at the FloWave tank \cite{flowave} where artificial waves and currents can be generated.
The tank is circumferentially ringed with 168 wavemakers, allowing for full control over the wave and current formation. To match the size of the vehicle a current of up to 0.5 m/s (~1 knot) and wave up to 10cm amplitude and 0.5Hz frequency were used. This was sufficient to introduce a significant disturbance in the vehicle's motion and can be used to represent how a much bigger offshore vehicle would behave in appropriately harsher conditions. This facility is capable of generating much stronger waves and currents, but the test volume is only 2m deep, so experiments on a larger scale with a bigger vehicle would not be feasible.

\section{Dataset}
\let\thefootnote\relax\footnotetext{Available for download at: \\ https://www.dropbox.com/sh/q8jyr6ays01s0x8/ \\ AACGAEHb8PHW33OpS3egHAt1a?dl=0}
 \begin{figure}[h]
 	\centering
 	\includegraphics[width=.65\linewidth]{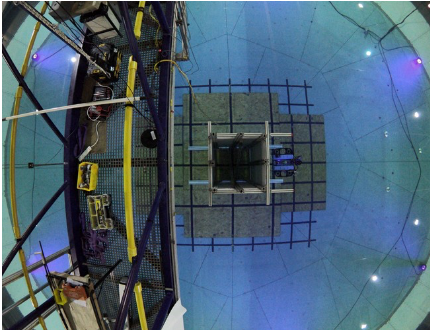}
 	\caption{The experimental tank and the structure that was inspected during the data collection. }
 	\label{fig:flowave}
 \end{figure} 

\begin{figure}[h]
	\centering
	\includegraphics[width=.65\linewidth]{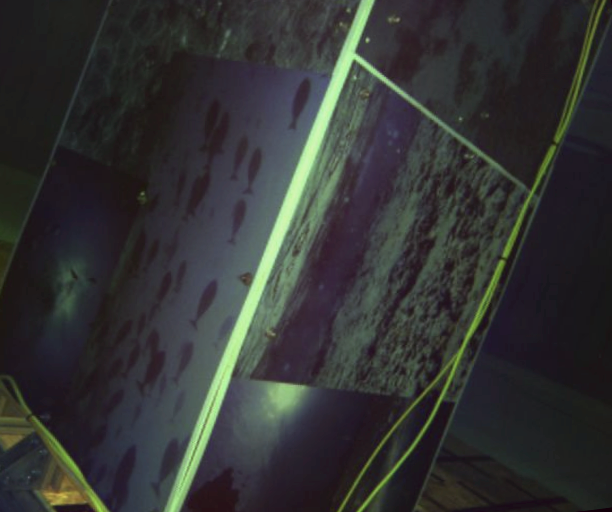}
	\caption{One of the images from the dataset with the view at the structure being inspected. }
	\label{fig:structure}
\end{figure}

The goal of the dataset prepared and recorded in this experiment, was to support the development of the navigation and SLAM algorithms for offshore, underwater inspections, but also to support the development of the manipulation and other intervention algorithms. To this end, two sets of experiments and recordings were performed. The first one, presented in Section \textit{Stereo odometry data} comprised the stereo images and the ground truth movement of the vehicle. The other (Section \textit{Movement disturbances for underwater interventions}), includes just the disruption to the vehicle's motion. During the data collection, an artificial structure was placed in the middle of the tank to be inspected by the ROV. It was built by attaching panels covered with high-resolution images of the underwater scenes to the cuboid frame (see Fig. \ref{fig:flowave} and \ref{fig:structure}). 

\subsection{Stereo odometry data}
\label{sec:setereo_odom_data}
To allow for the evaluation of the vision-based navigation and SLAM algorithms, multiple image sequences, under various conditions, were recorded. The first sequence is a long recording (5min) of the vehicle moving around the structure, with no waves or current (Figure \ref{fig:paths}, sequence $\#01$). This can be used as a baseline for the algorithm's performance in optimal conditions. The path of the vehicle is overlapping with itself in multiple places to allow for loop closures, place recognition tests etc. This is followed by 13 other recordings, where current and/or waves of increasing strength were introduced. These recordings are shorter (1 min each) and the vehicle was tasked to move back and forth in front of the structure. This is consistent with the task of approaching a given location at the offshore structure in increasingly difficult conditions. For currents above 0.3 m/s, no waves were used, as the vehicle was already getting unstable and data with additional disturbance would provide little value. For lower currents, waves 5cm/0.5Hz, 7.5cm/0.5Hz and 10cm/0.5Hz were used. Similar conditions with 1Hz wave frequency were also tested, but this seemed to have a smaller effect on the vehicle's motion and was ignored. The path of the vehicle, as obtained by the Qualysis system, during the provided experiments are visualised in Figure \ref{fig:paths}. 

\begin{figure}[h]
 	\centering
 	\includegraphics[width=0.45\linewidth]{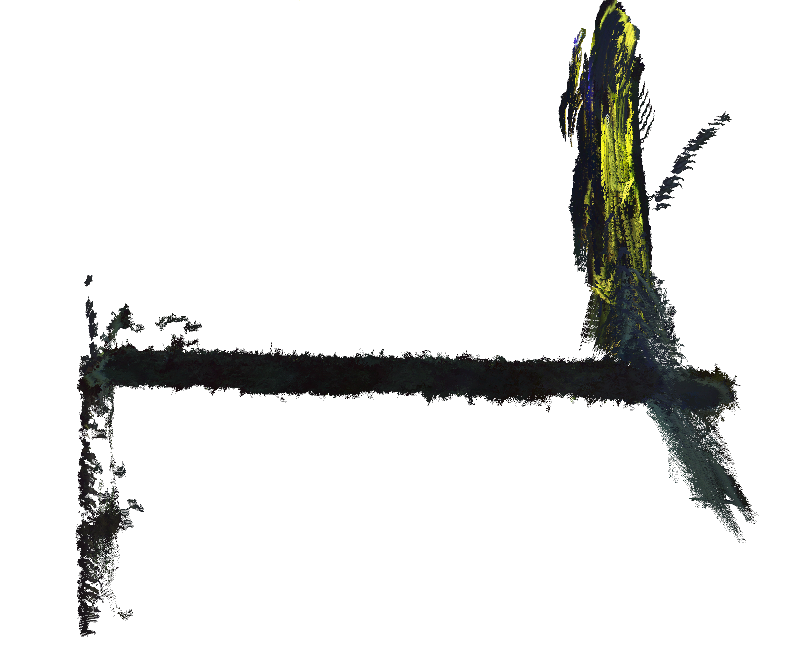}
 	\includegraphics[width=0.45\linewidth]{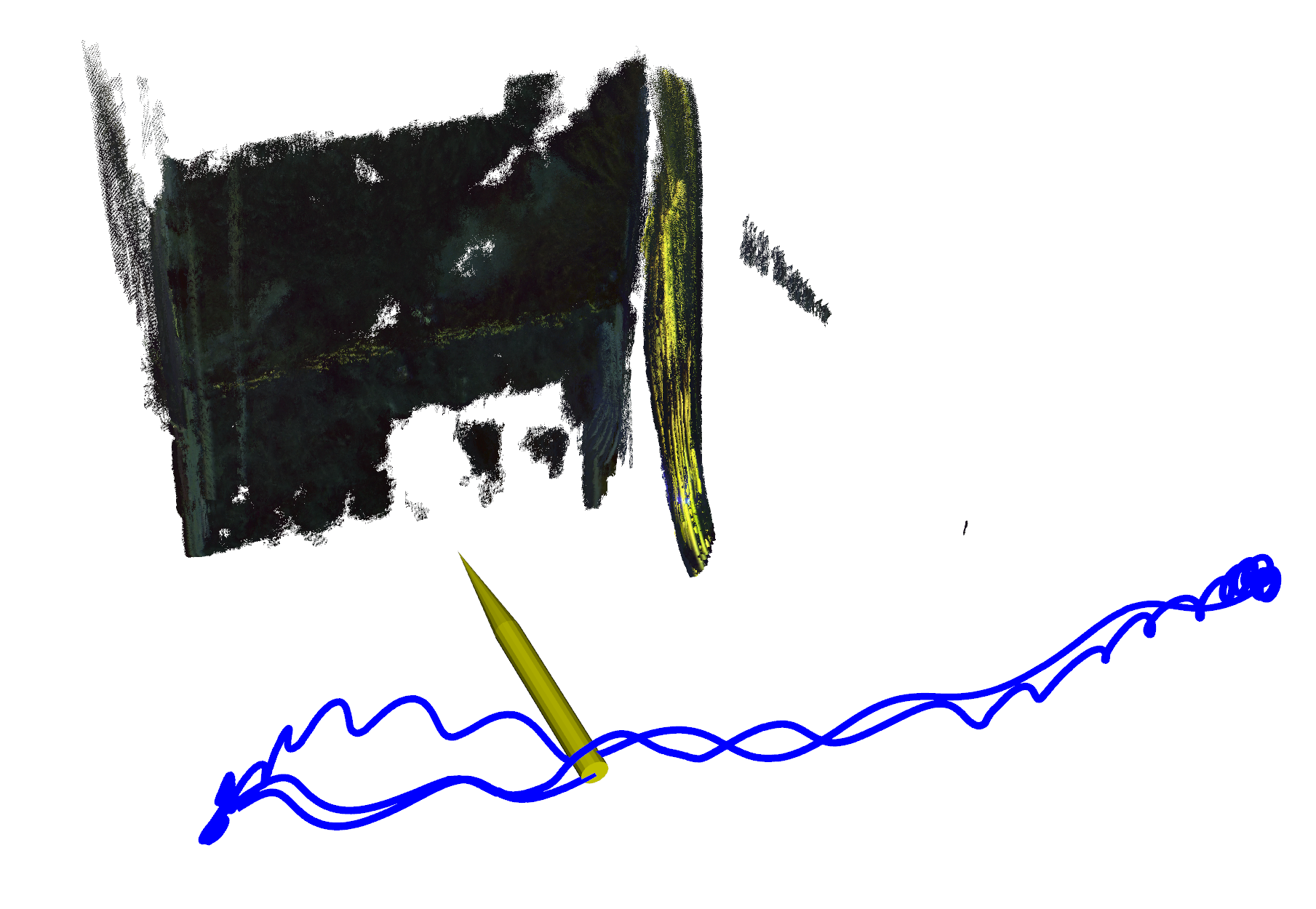}
 	\caption{Reconstruction of the structure with the accumulated pointcloud. Left: view from the bottom, Right: isometric perspective. Please note that good pointclouds' alignment and no warping indicates good system calibration. The yellow tether was free to move,causing some noise in the reconstruction. }
 	\label{fig:pc_asembled}
 \end{figure}

To validate the calibration of each element in the system and the quality of tracking, the structure used in the experiment was reconstructed by calculating 3D pointclouds for each consecutive stereo pair and accumulating them according to the pose from the Qualisys. The results (view from the bottom and the isometric perspective) are presented in the Figure \ref{fig:pc_asembled}. No additional filtering or alignment was done. All pointclouds seem to be aligned well, which indicates teat the motion tracking worked properly, and the calibration between the tracked and the camera frame was successful. Furthermore, pointclouds of the planar surfaces show no signs of warping or bending, indicating proper image rectification. The reconstruction around the tether is a bit noisy, but this is to be expected, as the tether was moving with the vehicle.

\subsection{Movement disturbances for underwater interventions}
\label{sec:movement}
One of the challenges for marine robotics is designing systems, that can reliably interact with underwater structures, especially under harsh conditions where the vehicle is unstable and the manipulator needs to compensate for the vehicle movement. This kind of systems must be extensively evaluated experimentally. However, the experimental setup used to create this dataset allowed for recording additional data samples that could be used in the development of intervention systems. 
The vehicle was station-keeping in the current and waves. This resulted in a slight motion caused by the disturbances that couldn't be fully compensated by the vehicle controller. The information about the vehicle's motion was recorded and processed. Firstly, the average pose was subtracted, to eliminate the offset. The data samples were also clipped such that the last pose is very close to the first one. This resulted in the translation and rotation offset, that can be used to simulate real disturbances. Clipping of the data also allows for using these samples in the loop. 

 \begin{figure}[h]
	\centering
	\includegraphics[width=.8\linewidth]{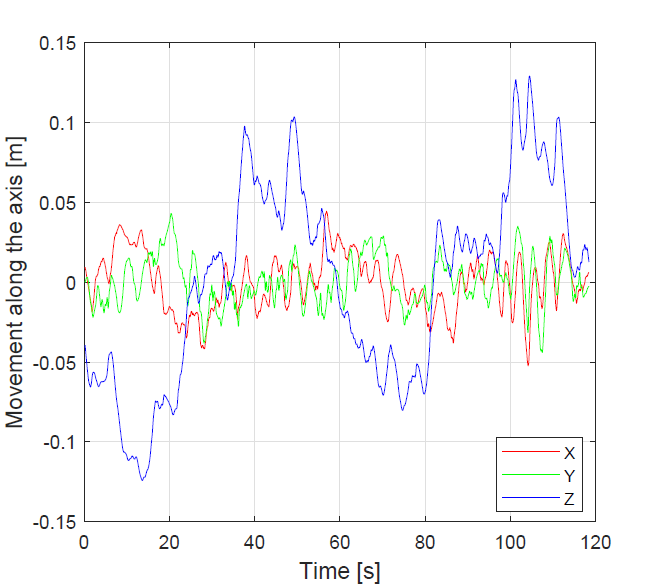}
	\caption{Movement of the vehicle caused by the disturbance: current 0.2m/s.}
	\label{fig:xyz1}
\end{figure} 

\begin{figure}[h]
	\centering
	\includegraphics[width=.8\linewidth]{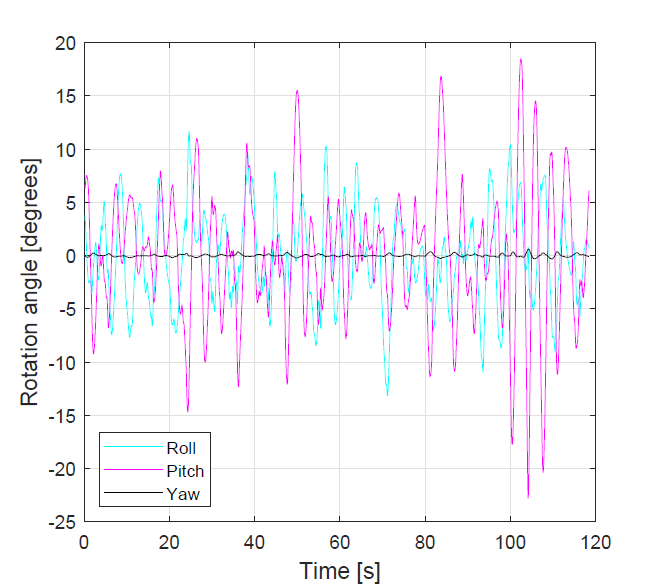}
	\caption{Rotation of the vehicle caused by the disturbance: current 0.2m/s.}
	\label{fig:rpy1}
\end{figure} 

\begin{figure}[h]
	\centering
	\includegraphics[width=.6\linewidth]{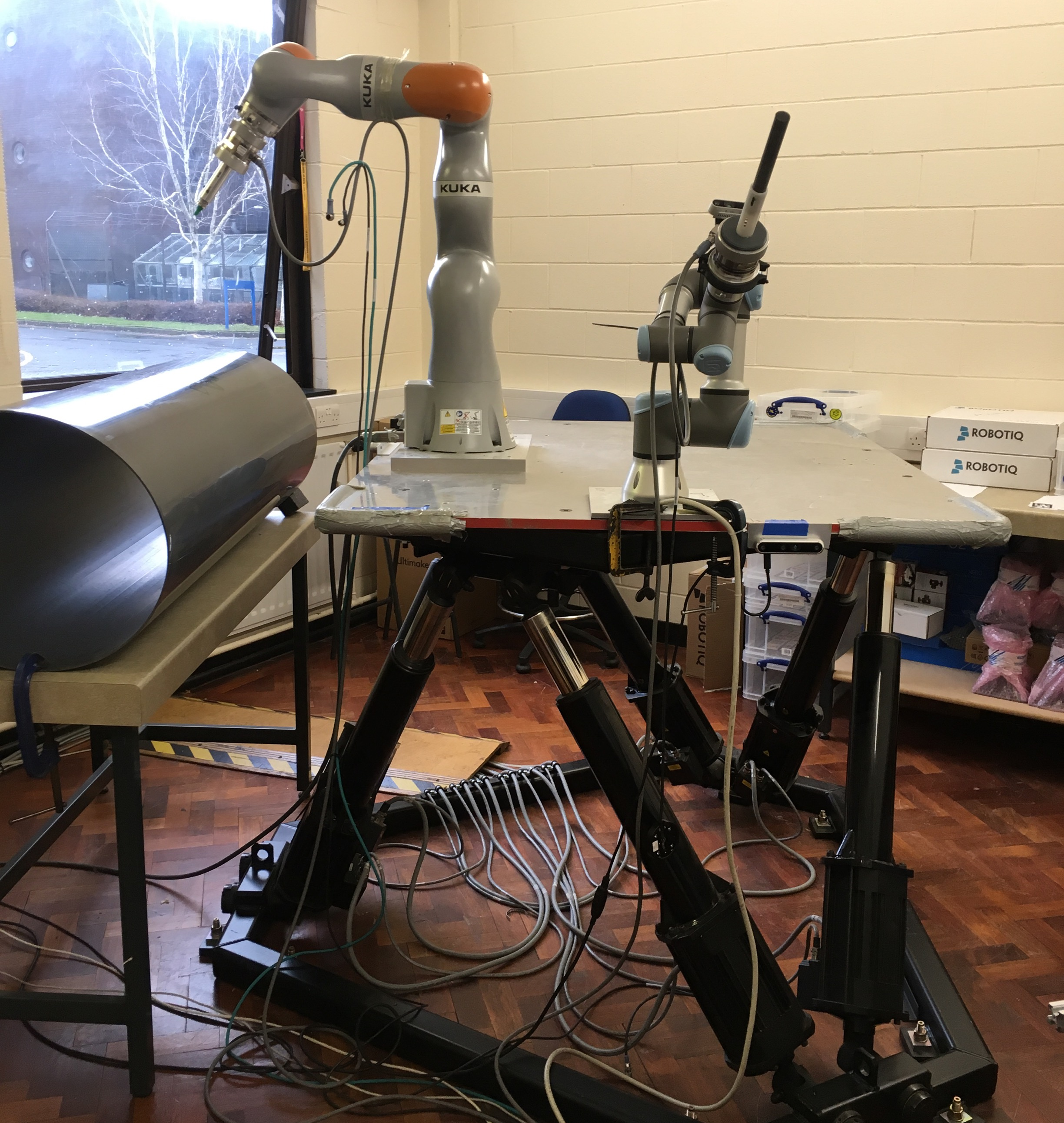}
	\caption{Example of the setup comprising a manipulator and a Stewart platform to simulate real motion of the ROV for the manipulation experiments.}
	\label{fig:sewart_platform}
\end{figure}

These disturbances can be used in simulations. Adding them to the simulated pose of the vehicle will result in the vehicle moving as programmed, but also being subjected to realistic disturbances. Furthermore, it is also possible to use this data in the manipulation experiments on real hardware. Figure \ref{fig:sewart_platform} shows the setup used in our experiments. The manipulator is attached to the Stewart platform. Disturbances, as present in this dataset, are used to control the Stewart platform. The manipulator's controller is not aware of this movement and is tasked with various manipulation experiments, where the arm needs to comply with the platform's movement.

\begin{figure}[h]
	\centering
	\includegraphics[width=.8\linewidth]{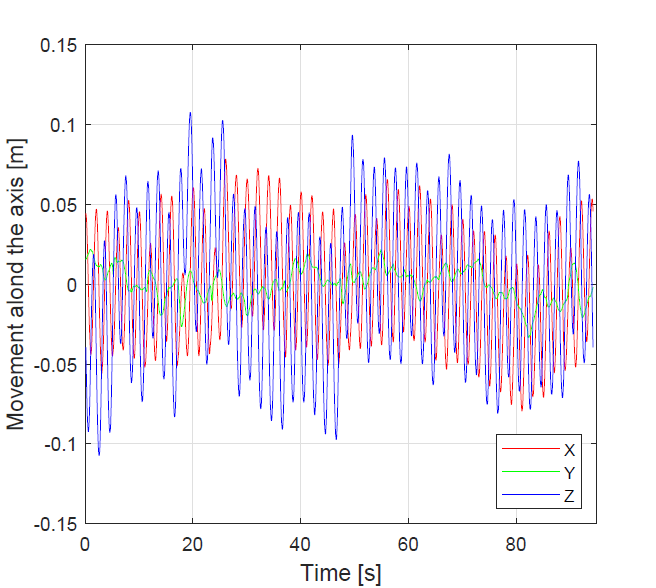}
	\caption{Movement of the vehicle caused by the disturbance: current 0.2m/s and wave: 10cm/0.5Hz.}
	\label{fig:xyz2}
\end{figure} 

\begin{figure}[h]
	\centering
	\includegraphics[width=.8\linewidth]{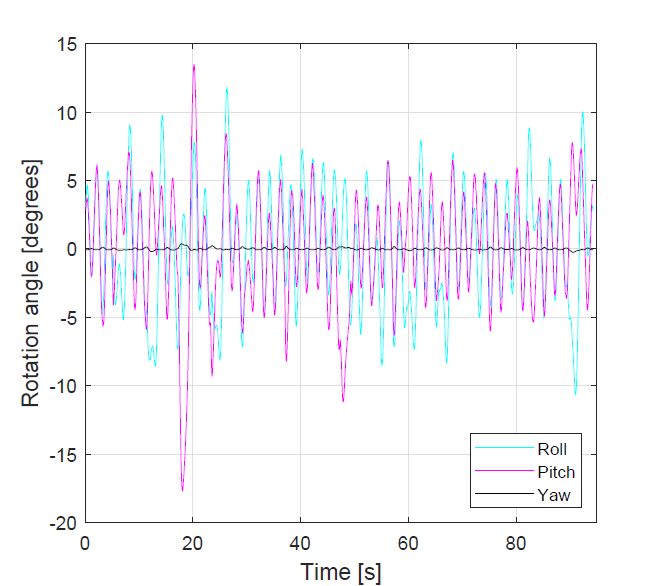}
	\caption{Rotation of the vehicle caused by the disturbance: current 0.2m/s and wave: 10cm/0.5Hz.}
	\label{fig:rpy2}
\end{figure} 

There are two recordings with movement disturbances included in the dataset. First one was collected in the presence of current, but without waves. This is presented in Figures \ref{fig:xyz1} and \ref{fig:rpy1}. The second sample was recorded with both waves and current (Figures \ref{fig:xyz2} and \ref{fig:rpy2}). Varying the current strength and wave parameters seemed to have been suppressed by the station keeping controller, only the frequency of the movement changed, the amplitude stayed the same, so only these two samples are included in the dataset.   

\section{Discussion}

\subsection{Data directory structure}
\begin{figure}[h]
	\centering
	\includegraphics[width=.6\linewidth]{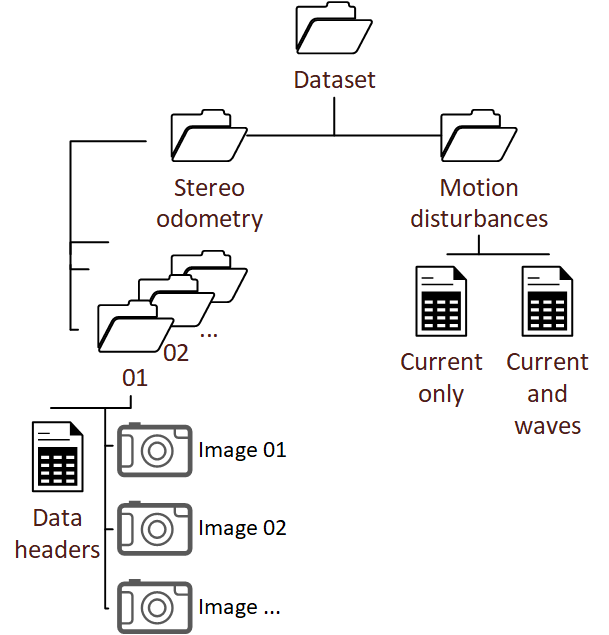}
	\caption{Dataset directory structure.}
	\label{fig:dir_structure}
\end{figure} 

The dataset is saved in the human-readable format, images are stored in the PNG files and all other information, including the corresponding timestamps, is stored in the CSV files. The data directory branches into the stereo dataset, as described in Section \textit{Stereo odometry data} and the motion disturbances recordings as discussed in Section \textit{Movement disturbances for underwater interventions}. 
Motion disturbances are fully contained in the CSV files, storing the timestamp, translation offset along the X, Y and Z axes as well as the roll, pitch and yaw. This data is also visualized on Figures \ref{fig:xyz1}, \ref{fig:rpy1}, \ref{fig:xyz2}, \ref{fig:rpy2}.
The stereo data directory branches further, storing 14 different data sequences, as presented in Figure \ref{fig:paths}. Each directory's name coincides with the sequence number as indicated here. Within these data samples' directories, sequences are summarized in the CSV file, storing timestamps corresponding to the images and the ground truth motion data. For each record, where an image is required, CSV stores the name of the image file. All images are saved in the same directory as the CSV file.

\subsection{Data parsing}

Parsing the data stored in the CSV format is fairly easy and can be implemented in any language. Along with the dataset, a Python script is provided to generate ROS-bags \cite{ROS} from the data. 


\subsection{Difficulties and shortcomings}

Despite our best effort to keep the data collected as consistent and easy to work with as possible, there are minor variations that should be mentioned. The lights used on the vehicle were impacting the quality of the motion tracking provided by Qualisys when the vehicle was moving rapidly. Therefore affected recordings were repeated with the lights off. The inspected structure can still easily be reconstructed, but the images from those recordings are visibly darker. Furthermore, in some places, a tether used to control the vehicle is captured in the field of view. It was moving slightly from frame to frame but should not confuse any state of the art visual odometry, as there was a comparatively big and static structure behind it. However, it can lead to some noise in the 3D reconstruction. This effect can be seen in Figure \ref{fig:pc_asembled}.

\section*{Acknowledgements}
The research presented here was undertaken under the EPSRC ORCA Hub project EP/R026173/1 (https://orcahub.org). The authors would like to thank Dr Roman Gabl and other members of the FloWave staff for their help in running the experiments.

\bibliographystyle{IEEEtran}
\bibliography{./main}

\end{document}